\documentclass[runningheads,a4paper]{llncs}

\usepackage{amssymb}
\usepackage{graphicx}

\usepackage{caption}
\usepackage{subcaption}
\usepackage{siunitx}
\DeclareSIUnit\perpixel{/pixel}

\usepackage{hyperref}
\urldef{\mailsa}\path|M.Veta@tue.nl|    
\urldef{\mailsb}\path|P.J.vanDiest@umcutrecht.nl|    
\urldef{\websitea}\path|http://tue.nl/image|    
\newcommand{\keywords}[1]{\par\addvspace\baselineskip
\noindent\keywordname\enspace\ignorespaces#1}

\usepackage{color}

\begin{document}

\mainmatter  

\title{Cutting out the middleman: measuring nuclear area in histopathology slides without segmentation}

\titlerunning{Measuring nuclear area in histopathology slides without segmentation}

%
%
\author{Mitko Veta \inst{1} \and Paul J. van Diest \inst{2} \and Josien P.W. Pluim \inst{1}}


\institute{Medical Image Analysis Group (IMAG/e), TU/e, Eindhoven,  The Netherlands \mailsa, \websitea \and
Department of Pathology, UMCU, Utrecht, The Netherlands}

%
%

\toctitle{Lecture Notes in Computer Science}
\tocauthor{Mitko Veta}
\maketitle

\begin{abstract}
The size of nuclei in histological preparations from excised breast tumors is predictive of patient outcome (large nuclei indicate poor outcome). Pathologists take into account nuclear size when performing breast cancer grading. In addition, the mean nuclear area (MNA) has been shown to have independent prognostic value. The straightforward approach to measuring nuclear size is by performing nuclei segmentation.  We hypothesize that given an image of a tumor region with known nuclei locations, the area of the individual nuclei and region statistics such as the MNA can be reliably computed directly from the image data by employing a machine learning model, without the intermediate step of nuclei segmentation. Towards this goal, we train a deep convolutional neural network  model that is applied locally at each nucleus location, and can reliably measure the area of the individual nuclei and the MNA. Furthermore, we show how such an approach can be extended to perform combined nuclei detection and measurement, which is reminiscent of granulometry.

\end{abstract}

\section{Introduction}

Cancer causes changes in the tissue phenotype (tissue appearance) that can be observed in histological tissue preparations. Based on the characteristics of the cancer phenotype, patients can be stratified into groups with different expected outcomes (recurrence or survival). Such characteristics identified by pathologists are arranged in grading systems. One such instance is the Bloom-Richardson grading system that is used for estimating the prognosis of breast cancer patients after surgical removal of the tumor. It consists of estimation of three biomarkers: nuclear pleomorhism, nuclear proliferation and tubule formation. Although such grading systems are routinely applied in clinical practice, they are known to suffer from reproducibility issues due to the subjectivity of the assessment. This can result in suboptimal estimation of the prognosis of the patient and in turn poor treatment planning. With the advent of digital pathology, which involves digitization of histological slides in the form of large, gigapixel images, automated quantitative image analysis is being proposed as a solution for this problem \cite{veta_breast_2014}.

In this paper we address the problem of measuring nuclear size in digitized histological slides from breast cancer patients. Estimation of the average nuclear size in the tissue by pathologists is part of the nuclear pleomorhism scoring (high grade tumors are characterizes by large nuclear size). In addition to being part of the Bloom-Richardson grading system, nuclear size expressed as the mean nuclear area (MNA) is an independent biomarker both by manual \cite{kronqvist_morphometric_1998,mommers_prognostic_2001} and automatic \cite{veta_prognostic_2012} measurement.

The straightforward approach to measuring the MNA of a tumor region is to detect the locations of all nuclei or of a representative sample, measure their area by segmentation and compute the average over the sample. Thus, when designing an automatic measurement method, the more general and difficult task of automatic nuclei segmentation needs to be solved first. We hypothesize that given an image of a tumor region with known nuclei locations, the area of the individual nuclei and region statistics such as the MNA can be reliably computed directly from the image data by employing a machine learning model, without the intermediate step of nuclei segmentation. With our approach, the machine learning model is applied locally and separately for each nucleus, i.e. on an image patch centered at the nucleus, and outputs an estimate of the nuclear area. Deep convolutional neural networks (CNN), that recently came into prominence and operate directly on raw image data are a natural fit for such an approach. This type of models has been successfully applied to a large variety of general computer vision tasks and are increasingly becoming relevant for medical image analysis \cite{ciresan_deep_2012,roth_new_2014,ginneken_off_2015} including histopathology image analysis \cite{ciresan_mitosis_2013}.  Additionally, we show how such an approach can be extended to perform combined nuclei detection and area measurement, without relying on manual input for the nuclei locations. This is reminiscent of granulometry \cite{vincent_fast_1994}, however, instead of using mathematical morphology operators, a  machine learning model that can better handle the complexity of histological images is used.

\section{Dataset}

The experiments in this paper were performed with the dataset of breast cancer histopathology images with manually segmented nuclei originally used in \cite{veta_automatic_2013}. This dataset consists of 39 slides from patients with invasive breast cancer. The slides were routinely prepared, stained  with hematoxylin and eosin (H\&E) and digitized at $\times 40$ magnification with a spatial resolution of \SI{0.25}{\micro\meter\perpixel}  at the Department of Pathology, University Medical Center Utrecht, The Netherlands. From each slide, representative tumor regions of size $1\times 1$ \SI{}{\milli\meter}  (resulting in images of size $4000\times4000$ pixels) were selected by an experienced pathologist. In each region, approximately 100 nuclei were selected with systematic random sampling and manually segmented by an expert observer (pathology resident). 

The dataset is divided in two subsets: subset A, consisting of 21 cases with 2191 manually segmented nuclei, is used as a training dataset, and subset B, consisting of 18 cases with 2073 manually segmented nuclei, is used as a testing dataset. For the experiments in this paper, subset A is further divided in two: subset A1 consisting of 14 cases, which is used for training the area measurement model, and subset A2 consisting of the remaining 7 cases, which is used as a validation dataset (to monitor for overfitting during training).

\section{Methods}

We first address the problem of measuring the area of nuclei with known locations (centroids) in the image using a machine learning model. Then, we show how such an approach can be extended to perform combined granulometry-like nuclei detection and area measurement.

\textbf{Area measurement as a classification task.} We assume that the locations of all or a sample of the nuclei in the image are known (we use the nuclei locations from the manual annotation by a pathologist). Given an image patch $\mathbf{x}$ centered at a nucleus location, we want to learn the parameters $\mathbf{w}$ of a function $f(\mathbf{x},\mathbf{w})$ that will approximate as closely as possible the area $y$ of the nucleus in the  center of the patch. This results to training a regression model.  However, we chose to treat this problem as classification. Instead of predicting the nuclear area directly with a regression model, a classification model can predict the bin of the area histogram to which a nucleus belongs (each histogram bin represents one class in the classification problem). The number of histogram bins defines the fidelity of the nuclei area measurement. The advantage of this over training a regression model is that it enables seamless extension to  a combined nuclei detection and area measurement model.

The output $f_c(\mathbf{x},\mathbf{w})$ of the classification model is a vector with probabilities associated with each class (area histogram bin). The area of the nucleus in $\mathbf{x}$ is reconstructed as the weighted average of the histogram bin centroids with the output probabilities used as weigths. This approach takes into account the confidence of the class prediction and results in a continuous output for the area measurements.

\textbf{Classification model.} We model $f_c(\mathbf{x}, \mathbf{w})$ as a deep CNN for classification. The deep CNN model consists of eight convolutional layers and two fully connected layers. As in  \cite{simonyan_very_2014}, we use filter size and padding combinations that preseve the input size, which simplifies the network design. The first convolutional layer has a kernel size of $5\times5$, and all remaining convolutional layers have kernels of size $3\times3$. The first, second, fifth and eighth layer are followed by a $2\times2$ max-pooling layer. The first two convolutional layers have 32 feature maps and the remaining six have 64 feature maps. The first fully connected layer has 128 neurons. The second fully connected layer (output layer) has a number of neurons equal to the number of classes and is followed by a softmax function. Dropout, which has a regularization effect, is applied after the last two max-pooling layers and between the two fully connected layers. Rectified linear unit (ReLU) nonlinearities are used throughout the network. 

\textbf{Training.} The range of nuclear areas in the training dataset, determined to be 16.6-\SI{151.8}{\square\micro\meter} based on the 0.5th and 99.5th percentile, was quantized into 20 histogram bins (20 classes for the classification problem). This number of histogram bins results in a reasonably small distance of \SI{6.8}{\square\micro\meter} between two neighboring bin centroids. Each nucleus was represented by a patch of size $96\times96$ pixels with a center corresponding to the nucleus centroid. This patch size is large enough to fit the largest nuclei in the dataset while still capturing some of the context. The number of classes and patch size were chosen based on optimization on the validation set, but the perfomance was stable for wide range of values.

\begin{figure}[t!]
\centering
\includegraphics[width=4cm,trim={0 0 0 .31cm},clip]{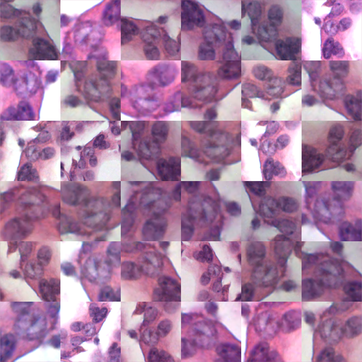}
\includegraphics[width=4cm,trim={0 0 0 .51cm},clip]{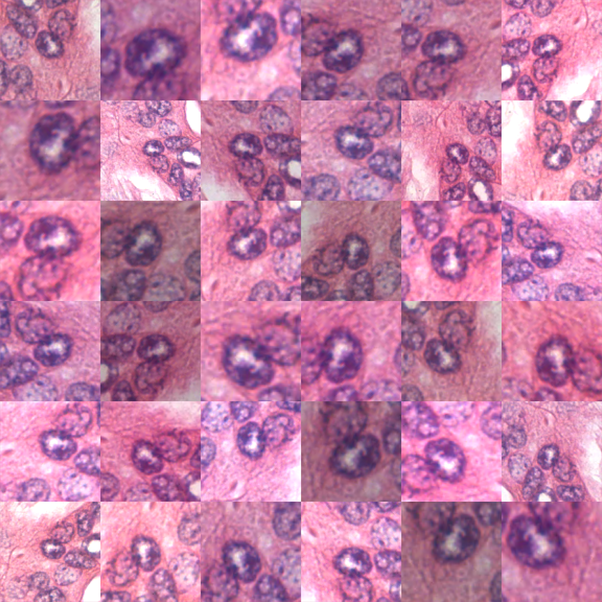}
\includegraphics[width=4cm,trim={0 0 0 .51cm},clip]{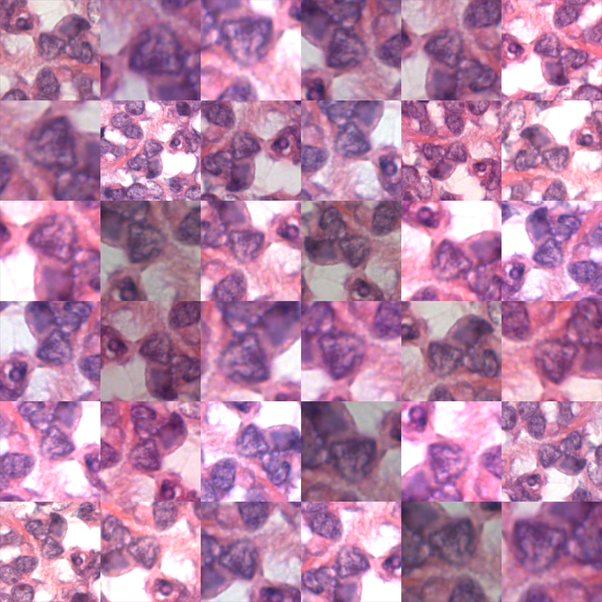}
\caption{Examples of data augmentation. New training samples are generated by random transformation of the original training samples.}
\label{augmentation}
\end{figure}

Since there is only a limited number of training samples in subset A1, data augmentation was necessary in order to avoid overfitting. The training samples were replicated by performing random translation, rotation, reflection, scaling, and color and contrast transformations. Note that the scaling transformation can change the class of the object, which is accounted for by changing the class label of the newly generated sample.  We used this property to balance the distribution of classes in the training set. Each nucleus in subset A1 was replicated 1000 times. This resulted in over 1.4 million training samples. Examples of the data augmentation are shown in Fig.~\ref{augmentation}. 

We used the Caffe \cite{jia_caffe_2014} deep learning framework to implement and optimize the deep CNN model. The model was optimized with batch gradient descent with batch size 256 and momentum 0.9. The base learning rate of 0.01 was decreased by 10\% of the current value every 2000 iterations. In addition to dropout and data augmentation, $L_2$ regularization was performed during training with weight decay value of 0.001. The weights of the neural network were initialized with small random numbers drawn from a uniform distribution. All biases were initialized to 0.1. The choice for these parameters was based on commonly used values for similar network configurations in the literature. The training was stopped after 25,000 iterations when the loss on the validation set (subset A2) stopped decreasing. 

\textbf{Combined nuclei detection and area measurement.} In order to train a model that can perform combined nuclei detection and classification, an additional ``background'' class is introduced in the classification task. This class accounts for patches that are not centered at nuclei locations. The classifier is then applied to every pixel location in a test image. The probability outputs for the ``background'' class are used to form a nuclei detection probability map. Local minima in this probability map below a certain threshold will correspond to nuclei centroids (the threshold value is the operating point of the detector and is subject to optimization). Once the nuclei are detected using the nuclei detection probability map, the same procedure as described before can be used to infer their size from the probability outputs of the ''foreground'' classes.

The annotated dataset used in this paper does not allow proper sampling of ``background'' patches for a training set, as only a limited number of the nuclei present in an image are annotated. In order to sample the ``background'' class, we used the results from the automatic segmentation method in \cite{veta_automatic_2013} as surrogate ground truth (we assume that the method correctly segments all nuclei in the image). The results from this method were used to create a mask of nuclei centroid locations. The ``background'' patches were then randomly sampled from the remaining image locations. Note that the surrogate ground truth was only used for sampling of the ``background'' class; the training samples for the remaining classes were based on the ground truth assigned by pathologists. 

From each image in subset A1 40,000 ``background'' patches were randomly sampled and together with the samples from the original 20 classes used to train the new classifier. We used a neural network architecture and a training procedure that is identical to the one described before. The training of this classification model converged after 40,000 iterations. For computational efficiency, the model was transformed to a fully convolutional neural network \cite{long_fully_2015} by converting the fully connected layers to convolutional layers.

\section{Experiments and Results}

We evaluate both nuclear area measurement with known nuclei locations and combined nuclei detection and area measurement. The former enables testing our hypothesis that nuclear area can be reliably measured by a machine learning model without performing segmentation under ideal conditions (manually annotated nuclei locations). 

The model for nuclear area measurement with known nuclei locations was trained with the manually annotated nuclei in subset A1, and then used to measure the nuclear area at the manually annotated nuclei locations in subset B. From the area measurements of the individual nuclei, the MNA was computed  for the 18 tumor regions in subset B. The measured area of individual nuclei and the MNA were compared with the measurements based on the manually segmented nuclei contours.

The combined nuclei detection and area measurement  model was trained with the manually annotated nuclei in subset A1 using the surrogate ground truth to sample the ``background class''. The optimal operating point of the detector was determined based on subset A2. The error of the estimation of the MNA over this subset was used as an optimization criterion. The trained model and the determined optimal operating point were then used to perform joint nuclei detection and area measurement in subset B. The resulting measurements were used to compute the MNA and compare it with the measurement based on the manually segmented nuclei contours.

The agreement between two sets of measurements was evaluated with the Bland-Altman method. In addition, the coefficient of determination for a linear fit between the two measurements was computed. 

\begin{figure}[t!]
    \centering
    \includegraphics[width=\textwidth]{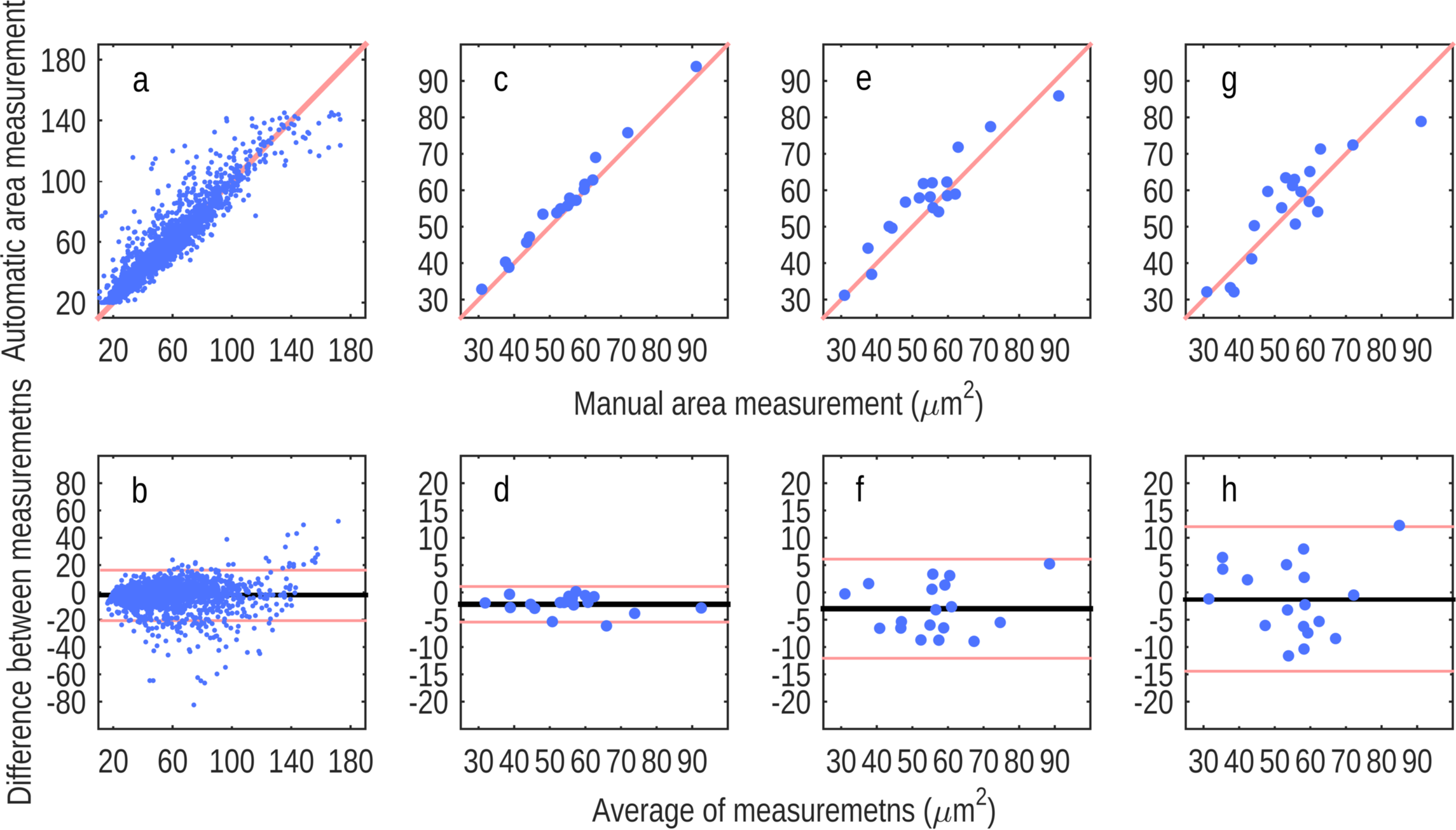}
    \caption{Scatter and Bland-Altman plots for manual and automatic measurement of nuclear area. (a) and (b) refer to the measurement of individual nuclei and (c) and (d) to the measurement of the MNA with the approach that relies on known nuclei locations. (e) and (f) refer to the measurement of the MNA with the combined nuclei detection and area measurement approach. (g) and (h) refer to the measurement of the MNA with the method described in \cite{veta_automatic_2013}. The red line in the scatter plots indicates the identity.}\label{main_results}
\end{figure}

\textbf{Nuclear area measurement with known nuclei locations.} The Bland-Altman plots and the corresponding scatter plots for the measurement of the area of individual nuclei and the MNA are shown in Fig.~\ref{main_results} (a-d). The  bias and limits of agreement for the measurement of the area of individual nuclei were $b = -2.19 \pm 18.85$ \SI{}{\square\micro\meter} and for the measurement of the MNA $b = -2.18 \pm 3.32$ \SI{}{\square\micro\meter}. The coefficient of determination was $ r^2 = 0.87$ for the measurement of the area of individual nuclei and  $r^2 = 0.99$ for the measurement of the MNA.

\begin{figure}[t!]
\centering
\includegraphics[width=2.95cm,trim={.266cm 0 0 .333cm},clip]{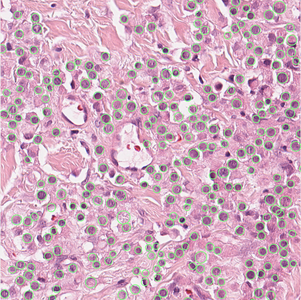}
\includegraphics[width=2.95cm,trim={.266cm 0 0 .333cm},clip]{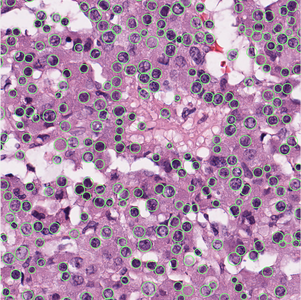}
\includegraphics[width=2.95cm,trim={.266cm 0 0 .333cm},clip]{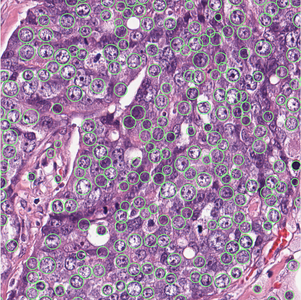}
\includegraphics[width=2.95cm,trim={.266cm 0 0 .333cm},clip]{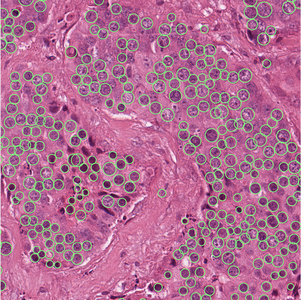}
\caption{Examples of combined nuclei detection and area measurement. The circles indicate the location and measured size of the nuclei (note that they are not countour segmentations). }
\label{granulometry}
\end{figure}

\textbf{Combined nuclei detection and area measurement.} The Bland-Altman plots and the corresponding scatter plots for the measurement of the MNA are shown in Fig.~\ref{main_results} (e, f). The bias and limits of agreement for the measurement of the MNA were $b = -2.98 \pm 9.26$ \SI{}{\square\micro\meter}. The coefficient of determination for the measurement of the MNA was $r^2 = 0.89$ . Some examples from the combined nuclei detection and area measurement are shown in Fig.~\ref{granulometry}.

\textbf{Comparison to measurement by automatic nuclei segmentation.}
For comparison, we show the results for the measurement of the MNA by performing nuclei segmentation with the method described in \cite{veta_automatic_2013}. The Bland-Altman plots and the corresponding scatter plots for the measurement of the MNA are shown in Fig.~\ref{main_results} (g, h). The bias and limits of agreement for the measurement of the MNA were $b = -1.20 \pm 13.50$ \SI{}{\square\micro\meter}. The coefficient of determination for the measurement of the MNA was $r^2 = 0.77$ .

\section{Discussion and Conclusions}

The Bland-Altman plot for the measurement of the area of individual nuclei with known locations (Fig.~\ref{main_results} (b)) indicates that there is a small bias in the automatic measurement. In other words, the nuclear area measured with the automatic method is on average larger when compared with the manual method. However, the bias value is very small considering the scale of nuclei sizes. The limits of agreement indicate moderate agreement with differences in the measurement that can be in the order of the area of the smallest nuclei in the dataset. Due to the averaging effect, the measurement of the MNA is considerably more accurate (Fig.~\ref{main_results} (d)). Although a small bias is still present, the limits of agreement indicate almost perfect agreement between the automatic and manual methods. This shows that the area of individual nuclei, and region statistics such as the MNA in particular, can be reliably computed directly from the image data without performing nuclei segmentation. These results, however, were achieved under ideal conditions, with expert annotations for the nuclei locations. The extension of this approach to combined nuclei detection and measurement has a much larger practical potential. The measurement of the MNA with this method had lower, but nevertheless substantial agreement with the manual measurement (Fig.~\ref{main_results} (f)). In part, the lower agreement is likely due to the two MNA measurements being based on different nuclei populations, although detection errors also have influence. This agreement was better compared with MNA measurement based on automatic nuclei segmentation (Fig.~\ref{main_results} (d)).

An added advantage of the methodology proposed in this paper is that deep CNNs can be efficiently run on GPUs. In our current implementation using fully convolutional neural networks, combined nuclei detection and area measurement in an image of size $4000\times4000$ pixels is performed in approximately 5 min. on a Tesla K40 GPU. We expect that this can be improved upon by exploiting the spatial redundancy of the image data (the current implementation evaluates the classifier at every pixel location), using smaller magnification such as $\times 20$, and optimizing the CNN architecture.

In future work we plan to use this methodology for automatic assesment of the prognosis of breast cancer patients. 

\bibliographystyle{splncs03}
\bibliography{bibliography/bibliography}

\end{document}